\documentclass[runningheads]{llncs}
\usepackage{graphicx}
\usepackage{amsmath}
\usepackage{amssymb}
\usepackage{tikz}
\usepackage{booktabs}
\newcommand{\keypoint}[1]{\vspace{0.1cm}\noindent\textbf{#1}\quad}

\begin{document}
\title{Hypernetwork Knowledge Graph Embeddings}
\author{Ivana Bala\v{z}evi\'c$^1$ \and
Carl Allen$^1$ \and
Timothy M. Hospedales$^{1,2}$}
\authorrunning{Bala\v{z}evi\'c et al.}
\institute{$^1$ School of Informatics, University of Edinburgh, UK\\
$^2$ Samsung AI Centre, Cambridge, UK\\
\email{\{ivana.balazevic, carl.allen, t.hospedales\}@ed.ac.uk}}
\maketitle              % typeset the header of the contribution
\begin{abstract}
Knowledge graphs are graphical representations of large data- bases of facts, which typically suffer from incompleteness. Inferring missing relations (links) between entities (nodes) is the task of link prediction. A recent state-of-the-art approach to link prediction, ConvE, implements a convolutional neural network to extract features from concatenated subject and relation vectors. Whilst results are impressive, the method is unintuitive and poorly understood. We propose a hypernetwork architecture that generates simplified relation-specific convolutional filters that (i) outperforms ConvE and all previous approaches across standard datasets; and (ii) can be framed as tensor factorization and thus set within a well established family of factorization models for link prediction. We thus demonstrate that convolution simply offers a convenient computational means of introducing sparsity and parameter tying to find an effective trade-off between non-linear expressiveness and the number of parameters to learn.

% \keywords{Link prediction \and Knowledge graphs \and Tensor factorization \and Convolutional neural networks.}
\end{abstract}
\section{Introduction}

Knowledge graphs, such as WordNet, Freebase, and Google Knowledge Graph, are large graph-structured databases of facts, containing information in the form of triples $(e_1, r, e_2)$, with $e_1$ and $e_2$ representing subject and object entities and $r$ a relation between them. They are considered important information resources, used for a wide variety of tasks ranging from question answering to information retrieval and text summarization. One of the main challenges with existing knowledge graphs is their incompleteness: many of the links between entities in the graph are missing. This has inspired substantial work in the field of \emph{link prediction}, i.e. the task of inferring missing links in knowledge graphs.

Until recently, many approaches to link prediction have been based on different factorizations of a 3-moded binary tensor representation of the training triples \cite{nickel2011three,socher2013reasoning,yang2014embedding,trouillon2016complex}. Such approaches are shallow and linear, with limited expressiveness. However, attempts to increase expressiveness with additional fully connected layers and non-linearities often lead to overfitting \cite{nickel2011three,socher2013reasoning}. For this reason, Dettmers et al. introduce ConvE, a model that uses 2D convolutions over reshaped and concatenated entity and relation embeddings \cite{dettmers2017convolutional}. They motivate the use of convolutions by being parameter efficient and fast to compute on a GPU, as well as having various robust methods from computer vision to prevent overfitting. Even though results achieved by ConvE are impressive, it is highly unintuitive that convolution -- particularly 2D convolution -- should be effective for extracting information from 1D entity and relation embeddings.

In this paper, we introduce HypER, a model that uses a \emph{hypernetwork} \cite{ha2016hypernetworks} to generate convolutional filter weights for each relation. A hypernetwork is an approach by which one network generates weights for another network, that can be used to enable weight-sharing across layers and to dynamically synthesize weights given an input. In our context, we generate relation-specific filter weights to process input entities, and also achieve \emph{multi-task knowledge sharing} across relations in the knowledge graph. Our proposed HypER model uses a hypernetwork to generate a set of 1D relation-specific filters to process the subject entity embeddings. This simplifies the interaction between subject entity and relation embeddings compared to ConvE, in which a global set of \emph{2D} filters are convolved over \emph{reshaped and concatenated} subject entity and relation embeddings, which is unintuitive as it suggests the presence of 2D structure in word embeddings. Moreover, interaction between subject and relation in ConvE depends on an arbitrary choice about how they are reshaped and concatenated. In contrast, HypER's hypernetwork generates relation-specific filters, and thus extracts \emph{relation-specific features} from the subject entity embedding. This necessitates no 2D reshaping, and allows entity and relation to interact more completely, rather than only around the concatenation boundary. We show that this simplified approach, in addition to improving link prediction performance, can be understood in terms of tensor factorization, thus placing HypER within a well established family of factorization models. The apparent obscurity of using convolution within word embeddings is thereby explained as simply a convenient computational means of introducing \emph{sparsity} and \emph{parameter tying}.

We evaluate HypER against several previously proposed link prediction models using standard datasets (FB15k-237, WN18RR, FB15k, WN18, YAGO3-10), across which it consistently achieves state-of-the-art performance. In summary, our key contributions are:
\begin{itemize}
\itemsep0em
\item proposing a new model for link prediction (HypER) which achieves state-of-the-art performance across all standard datasets;
\item showing that the benefit of using convolutional instead of fully connected layers is due to restricting the number of dimensions that interact (i.e. explicit regularization), rather than finding higher dimensional structure in the embeddings (as implied by ConvE); and
\item showing that HypER in fact falls within a broad class of tensor factorization models despite the use of convolution, which serves to provide a good trade-off between expressiveness and number of parameters to learn.
\end{itemize}

\section{Related Work}

Numerous matrix factorization approaches to link prediction have been proposed. An early model, RESCAL \cite{nickel2011three}, tackles the link prediction task by optimizing a scoring function containing a bilinear product between vectors for each of the subject and object entities and a full rank matrix for each relation. DistMult \cite{yang2014embedding} can be viewed as a special case of RESCAL with a diagonal matrix per relation type, which limits the linear transformation performed on entity vectors to a stretch. ComplEx \cite{trouillon2016complex} extends DistMult to the complex domain. TransE  \cite{bordes2013translating} is an affine model that represents a relation as a translation operation between subject and object entity vectors.

A somewhat separate line of link prediction research introduces Relational Graph Convolutional Networks (R-GCNs) \cite{schlichtkrull2018modeling}. R-GCNs use a convolution operator to capture locality information in graphs. The model closest to our own and which we draw inspiration from, is ConvE \cite{dettmers2017convolutional}, where a convolution operation is performed on the subject entity vector and the relation vector, after they are each reshaped to a matrix and lengthwise concatenated. The obtained feature maps are flattened, put through a fully connected layer, and the inner product is taken with all object entity vectors to generate a score for each triple. Advantages of ConvE over previous approaches include its expressiveness, achieved by using multiple layers of non-linear features, its scalability to large knowledge graphs, and its robustness to overfitting. However, it is not intuitive why convolving across concatenated and reshaped subject entity and relation vectors should be effective.

The proposed HypER model does no such reshaping or concatenation and thus avoids both implying any inherent 2D structure in the embeddings and restricting interaction to the concatenation boundary. Instead, HypER convolves \emph{every} dimension of the subject entity embedding with \emph{relation-specific} convolutional filters generated by the hypernetwork. This way, entity and relation embeddings are combined in a non-linear (quadratic) manner, unlike the linear combination (weighted sum) in ConvE.  This gives HypER more expressive power, while also reducing parameters. 

Interestingly, we find that the differences in moving from ConvE to HypER in fact bring the factorization and convolutional approaches together, since the 1D convolution process is equivalent to multiplication by a highly sparse tensor with tied weights (see Figure \ref{fig:hypermatrix}). The multiplication of this ``convolutional tensor'' (defined by the relation embedding and hypernetwork) and other weights gives an implicit relation matrix, corresponding to those in  e.g. RESCAL, DistMult and ComplEx. Other than the method of deriving these relation matrices, the key difference to existing factorization approaches is the ReLU non-linearity applied prior to interaction with the object embedding. 

\begin{table}[!t]
\centering
\caption{Scoring functions of state-of-the-art link prediction models, the dimensionality of their relation parameters, and their space complexity. $d_e$ and $d_r$ are the dimensions of entity and relation embeddings respectively, $\overline{\mathbf{e}}_2 \in \mathbb{C}^{d_e}$ denotes the complex conjugate of $\mathbf{e}_2$, and $\underline{\mathbf{e}}_1, \underline{\mathbf{w}}_r \in \mathbb{R}^{d_w\times d_h}$ denote a 2D reshaping of $\mathbf{e}_1$ and $\mathbf{w}_r$ respectively. $*$ is the convolution operator, $\mathbf{F}_r = \text{vec}^{-1}(\mathbf{w}_r\mathbf{H})$ the matrix of relation specific convolutional filters, $\text{vec}$ is a vectorization of a matrix and $\text{vec}^{-1}$ its inverse, $f$ is a non-linear function, and $n_e$ and $n_r$ respectively denote the number of entities and relations.
  }
\resizebox{11cm}{!}{
\begin{tabular}{lcccc}
  \toprule
  \textbf{Model} & \textbf{Scoring Function} & \textbf{Relation Parameters} & \textbf{Space Complexity}\\
  \midrule
   RESCAL \cite{nickel2011three} & $\mathbf{e}_1^\top\mathbf{W}_r\mathbf{e}_2$ & $\mathbf{W}_r \in \mathbb{R}^{{d_e}^2}$ & $\mathcal{O}(n_e d_e + n_r d_e^2)$ \\
%   \hline
   TransE \cite{bordes2013translating} & $\|\mathbf{e}_1 + \mathbf{w}_r - \mathbf{e}_2\|$ & $\mathbf{w}_r \in \mathbb{R}^{d_e}$ & $\mathcal{O}(n_e d_e + n_r d_e)$ \\
%   \hline
  NTN \cite{socher2013reasoning} & $\mathbf{u}_r^\top f(\mathbf{e}_1\mathbf{W}_r^{[1..k]}\mathbf{e}_2 + \mathbf{V}_r   \begin{bmatrix} \mathbf{e}_1 \\ \mathbf{e}_2\end{bmatrix}  + \mathbf{b}_r)$ & \parbox{4cm}{\centering$\mathbf{W}_r \in \mathbb{R}^{{d_e}^2k}, \mathbf{V}_r \in \mathbb{R}^{2d_e k},$\\ $\mathbf{u}_r \in \mathbb{R}^k, \mathbf{b}_r \in \mathbb{R}^k$} & $\mathcal{O}(n_e d_e + n_r {d_e}^2k)$ \\
% %   \hline
   DistMult \cite{yang2014embedding} & $\langle\mathbf{e}_1, \mathbf{w}_r, \mathbf{e}_2 \rangle$ & $\mathbf{w}_r \in \mathbb{R}^{d_e}$ & $\mathcal{O}(n_e d_e + n_r d_e)$ \\
%   \hline
   ComplEx \cite{trouillon2016complex} & $\text{Re}(\langle\mathbf{e}_1, \mathbf{w}_r, \overline{\mathbf{e}}_2 \rangle)$ & $\mathbf{w}_r \in \mathbb{C}^{d_e}$ & $\mathcal{O}(n_e d_e + n_r d_e)$ \\
%   \hline
   ConvE \cite{dettmers2017convolutional} & $f(\text{vec}(f([\underline{\mathbf{e}}_1; \underline{\mathbf{w}}_r] * w))\mathbf{W})\mathbf{e}_2$ & $\mathbf{w}_r \in \mathbb{R}^{d_r}$ & $\mathcal{O}(n_e d_e + n_r d_r)$\\
%   \hline
   HypER (ours) & $f(\text{vec}(\mathbf{e}_1 * \text{vec}^{-1}(\mathbf{w}_r\mathbf{H}))\mathbf{W})\mathbf{e}_2$ & $\mathbf{w}_r \in \mathbb{R}^{d_r}$ & $\mathcal{O}(n_e d_e + n_r d_r)$\\
  \bottomrule
\end{tabular}
}
  \label{table:models}
  \vspace{-0.5cm}
\end{table}

\vspace{-0.3cm}
\section{Link Prediction} \label{sec:link}

In link prediction, the aim is to learn a scoring function $\phi$ that assigns a score $s = \phi(e_1, r, e_2) \in \mathbb{R}$ to each input triple $(e_1, r, e_2)$, where $e_1, e_2 \in \mathcal{E}$ are subject and object entities and $r \in \mathcal{R}$ a relation. The score indicates the strength of prediction that the given triple corresponds to a true fact, with positive scores meaning true and negative scores, false. Link prediction models typically map entity pair $e_1, e_2$ to their corresponding distributed embedding representations $\mathbf{e}_1, \mathbf{e}_2 \in \mathbb{R}^{d_e}$ and a score is assigned using a \emph{relation-specific} function, $s = \phi_r(\mathbf{e}_1, \mathbf{e}_2)$. The majority of link prediction models apply the logistic sigmoid function $\sigma(\cdot)$ to the score to give a probabilistically interpretable prediction $p = \sigma(s) \in [0, 1]$ as to whether the queried fact is true. The scoring functions for models from across the literature and HypER are summarized in Table~\ref{table:models}, together with the dimensionality of their relation parameters and the significant terms of their space complexity.

\vspace{-0.3cm}
\section{Hypernetwork Knowledge Graph Embeddings}

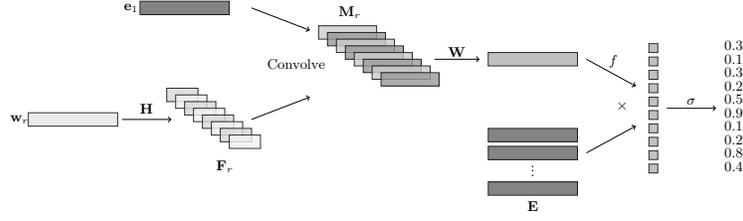
\begin{figure}[!ht]
\centering
\resizebox{10cm}{!}{
\begin{tikzpicture}
\node[black] at (2.3, 0.25, 0) {\small $\mathbf{e}_1$};
\draw[black,opacity=0.8,fill=black!60] (2.5, 0.1, 0) -- (4.5, 0.1, 0) -- (4.5, 0.4, 0) -- (2.5, 0.4, 0) -- cycle;% e1
\draw [->] (5, 0.25, 0) -- (6.3, -0.25, 0);
\node[black] at (7.2, 0.15, 0) {\small $\mathbf{M}_r$};
\draw[black,opacity=0.8,fill=black!20] (6.5, -0.45, 0) -- (7.8, -0.45, 0) -- (7.8, -0.15, 0) -- (6.5, -0.15, 0) -- cycle;
\draw[black,opacity=0.8,fill=black!40] (6.7, -0.6, 0) -- (8, -0.6, 0) -- (8, -0.3, 0) -- (6.7, -0.3, 0) -- cycle;
\draw[black,opacity=0.8,fill=black!20] (6.9, -0.75, 0) -- (8.2, -0.75, 0) -- (8.2, -0.45, 0) -- (6.9, -0.45, 0) -- cycle;
\draw[black,opacity=0.8,fill=black!40] (7.1, -0.9, 0) -- (8.4, -0.9, 0) -- (8.4, -0.6, 0) -- (7.1, -0.6, 0) -- cycle;
\draw[black,opacity=0.8,fill=black!20] (7.3, -1.05, 0) -- (8.6, -1.05, 0) -- (8.6, -0.75, 0) -- (7.3, -0.75, 0) -- cycle;
\draw[black,opacity=0.8,fill=black!40] (7.5, -1.2, 0) -- (8.8, -1.2, 0) -- (8.8, -0.9, 0) -- (7.5, -0.9, 0) -- cycle;
\draw[black,opacity=0.8,fill=black!20] (7.7, -1.35, 0) -- (9, -1.35, 0) -- (9, -1.05, 0) -- (7.7, -1.05, 0) -- cycle;
\draw[black,opacity=0.8,fill=black!40] (7.9, -1.5, 0) -- (9.2, -1.5, 0) -- (9.2, -1.2, 0) -- (7.9, -1.2, 0) -- cycle;
\draw [->] (9.1, -0.9, 0) -- (10.1, -0.9, 0);
\node[black] at (9.6, -0.75, 0) {\small $\mathbf{W}$};
\draw[black,opacity=0.8,fill=black!30] (10.3, -0.75, 0) -- (12.3, -0.75, 0) -- (12.3, -1.05, 0) -- (10.3, -1.05, 0) -- cycle;
\draw [->] (12.5, -0.9, 0) -- (13.6, -1.5, 0);
\node[black] at (13.1, -0.95, 0) {\small $f$};

\draw[black,opacity=0.8,fill=black!60] (10.3, -2.45, 0) -- (12.3, -2.45, 0) -- (12.3, -2.75, 0) -- (10.3, -2.75, 0) -- cycle;
\draw[black,opacity=0.8,fill=black!60] (10.3, -2.85, 0) -- (12.3, -2.85, 0) -- (12.3, -3.15, 0) -- (10.3, -3.15, 0) -- cycle;
\node[black] at (11.3, -3.3, 0) {.};
\node[black] at (11.3, -3.4, 0) {.};
\node[black] at (11.3, -3.5, 0) {.};
\draw[black,opacity=0.8,fill=black!60] (10.3, -3.65, 0) -- (12.3, -3.65, 0) -- (12.3, -3.95, 0) -- (10.3, -3.95, 0) -- cycle;
\node[black] at (11.3, -4.2, 0) {\small $\mathbf{E}$};
\draw [->] (12.5, -3., 0) -- (13.6, -2.4, 0);
\node[black] at (13.3, -1.95, 0) {$\times$};

\draw[black,opacity=0.8,fill=black!30] (13.9, -0.75, 0) -- (14.1, -0.75, 0) -- (14.1, -0.55, 0) -- (13.9, -0.55, 0) -- cycle;
\draw[black,opacity=0.8,fill=black!30] (13.9, -1.05, 0) -- (14.1, -1.05, 0) -- (14.1, -0.85, 0) -- (13.9, -0.85, 0) -- cycle;
\draw[black,opacity=0.8,fill=black!30] (13.9, -1.35, 0) -- (14.1, -1.35, 0) -- (14.1, -1.15, 0) -- (13.9, -1.15, 0) -- cycle;
\draw[black,opacity=0.8,fill=black!30] (13.9, -1.65, 0) -- (14.1, -1.65, 0) -- (14.1, -1.45, 0) -- (13.9, -1.45, 0) -- cycle;
\draw[black,opacity=0.8,fill=black!30] (13.9, -1.95, 0) -- (14.1, -1.95, 0) -- (14.1, -1.75, 0) -- (13.9, -1.75, 0) -- cycle;
\draw[black,opacity=0.8,fill=black!30] (13.9, -2.25, 0) -- (14.1, -2.25, 0) -- (14.1, -2.05, 0) -- (13.9, -2.05, 0) -- cycle;
\draw[black,opacity=0.8,fill=black!30] (13.9, -2.55, 0) -- (14.1, -2.55, 0) -- (14.1, -2.35, 0) -- (13.9, -2.35, 0) -- cycle;
\draw[black,opacity=0.8,fill=black!30] (13.9, -2.85, 0) -- (14.1, -2.85, 0) -- (14.1, -2.65, 0) -- (13.9, -2.65, 0) -- cycle;
\draw[black,opacity=0.8,fill=black!30] (13.9, -3.15, 0) -- (14.1, -3.15, 0) -- (14.1, -2.95, 0) -- (13.9, -2.95, 0) -- cycle;
\draw[black,opacity=0.8,fill=black!30] (13.9, -3.45, 0) -- (14.1, -3.45, 0) -- (14.1, -3.25, 0) -- (13.9, -3.25, 0) -- cycle;

\node[black] at (4.4, -3.3, 0) {\small $\mathbf{F}_r$};
\draw [->] (14.3, -2, 0) -- (15.4, -2, 0);
\node[black] at (14.85, -1.85, 0) {\small $\sigma$};

\node[black] at (15.8, -0.6, 0) {\small 0.3};
\node[black] at (15.8, -0.9, 0) {\small 0.1};
\node[black] at (15.8, -1.2, 0) {\small 0.3};
\node[black] at (15.8, -1.5, 0) {\small 0.2};
\node[black] at (15.8, -1.8, 0) {\small 0.5};
\node[black] at (15.8, -2.1, 0) {\small 0.9};
\node[black] at (15.8, -2.4, 0) {\small 0.1};
\node[black] at (15.8, -2.7, 0) {\small 0.2};
\node[black] at (15.8, -3, 0) {\small 0.8};
\node[black] at (15.8, -3.3, 0) {\small 0.4};

\node[black] at (-0.2, -2.25, 0) {\small $\mathbf{w}_r$};
\draw[black,opacity=0.8,fill=black!10] (0, -2.1, 0) -- (2, -2.1, 0) -- (2, -2.4, 0) -- (0, -2.4, 0) -- cycle;% r
\draw [->] (2.1, -2.25, 0) -- (3.2, -2.25, 0);
\node[black] at (2.65, -2, 0) {\small $\mathbf{H}$};
% \node[black] at (2.65, -2.5, 0) {\small reshape};
\draw[black,opacity=0.8,fill=black!15] (3.1, -1.85, 0) -- (3.8, -1.85, 0) -- (3.8, -1.55, 0) -- (3.1, -1.55, 0) -- cycle;
\draw[black,opacity=0.8,fill=black!5] (3.3, -2., 0) -- (4, -2., 0) -- (4, -1.7, 0) -- (3.3, -1.7, 0) -- cycle;
\draw[black,opacity=0.8,fill=black!15] (3.5, -2.15, 0) -- (4.2, -2.15, 0) -- (4.2, -1.85, 0) -- (3.5, -1.85, 0) -- cycle;
\draw[black,opacity=0.8,fill=black!5] (3.7, -2.3, 0) -- (4.4, -2.3, 0) -- (4.4, -2, 0) -- (3.7, -2, 0) -- cycle;
\draw[black,opacity=0.8,fill=black!15] (3.9, -2.45, 0) -- (4.6, -2.45, 0) -- (4.6, -2.15, 0) -- (3.9, -2.15, 0) -- cycle;
\draw[black,opacity=0.8,fill=black!5] (4.1, -2.6, 0) -- (4.8, -2.6, 0) -- (4.8, -2.3, 0) -- (4.1, -2.3, 0) -- cycle;
\draw[black,opacity=0.8,fill=black!15] (4.3, -2.75, 0) -- (5, -2.75, 0) -- (5, -2.45, 0) -- (4.3, -2.45, 0) -- cycle;
\draw[black,opacity=0.8,fill=black!5] (4.5, -2.9, 0) -- (5.2, -2.9, 0) -- (5.2, -2.6, 0) -- (4.5, -2.6, 0) -- cycle;
\draw [->] (5, -2.25, 0) -- (6.3, -1.75, 0);
\node[black] at (6, -1., 0) {\small Convolve};
\end{tikzpicture}
}
\caption{Visualization of the HypER model architecture. Subject entity embedding $\mathbf{e}_1$ is convolved with filters $\mathbf{F}_r$, created by the hypernetwork $\mathbf{H}$ from relation embedding $\mathbf{w}_r$. The obtained feature maps $\mathbf{M}_r$ are mapped to $d_e$-dimensional space via $\mathbf{W}$ and the non-linearity $f$ applied before being combined with all object vectors $\mathbf{e}_2\in \mathbf{E}$ through an inner product to give a score for each triple. Predictions are obtained by applying the logistic sigmoid function to each score.}
\label{fig:hyper}
\end{figure}

In this work, we propose a novel hypernetwork model for link prediction in knowledge graphs. In summary, the hypernetwork projects a vector embedding of each relation via a fully connected layer, the result of which is reshaped to give a set of convolutional filter weight vectors for each relation. We explain this process in more detail below. The idea of using convolutions on entity and relation embeddings stems from computer vision, where feature maps reflect patterns in the image such as lines or edges. Their role in the text domain is harder to interpret, since little is known of the meaning of a single dimension in a word embedding. We believe convolutional filters have a regularizing effect when applied to word embeddings (compared to the corresponding full tensor), as the filter size restricts which dimensions of embeddings can interact. This allows nonlinear expressiveness while limiting overfitting by using few parameters. A visualization of  HypER  is given in Figure \ref{fig:hyper}.

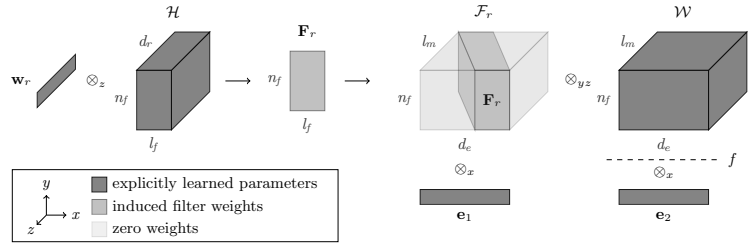
\begin{figure}[!ht]
  \centering
  \resizebox{10cm}{!}{
  \begin{tikzpicture}
    \pgfmathsetmacro{\cubex}{0.7}
    \pgfmathsetmacro{\cubey}{1.2}
    \pgfmathsetmacro{\cubez}{2}

	\node[black] at (2.3, 0, 0) {$\mathbf{w}_r$};
    \draw[black,opacity=0.8,fill=black!60] (3, -0.15, -1) -- ++(0, 0, +\cubez) -- ++(0, 0.3, 0) -- ++(0, 0, -\cubez) -- cycle;% e1
    \node[black] at (3.8, 0, 0) {$\mathbf{\otimes}_z$};

	% hypernetwork
    \pgfmathsetmacro{\cubexp}{5.7}
    \pgfmathsetmacro{\cubeyp}{\cubey/2}
    \pgfmathsetmacro{\cubezp}{\cubez/2}
    \node[black] at (\cubexp-\cubex/2, \cubeyp+\cubey/2+0.2, \cubezp-\cubez/2) {$\mathcal{H}$};
    \draw[black,opacity=0.8,fill=black!60] (\cubexp,\cubeyp,\cubezp) -- ++(-\cubex,0,0) -- ++(0,-\cubey,0) -- ++(\cubex,0,0) -- cycle;
    \draw[black,opacity=0.8,fill=black!60] (\cubexp,\cubeyp,\cubezp) -- ++(0,0,-\cubez) -- ++(0,-\cubey,0) -- ++(0,0,\cubez) -- cycle;
    \draw[black,opacity=0.8,fill=black!60] (\cubexp,\cubeyp,\cubezp) -- ++(-\cubex,0,0) -- ++(0,0,-\cubez) -- ++(\cubex,0,0) -- cycle;
    \node[black!80] at (\cubexp-\cubex-0.2, \cubeyp+0.2, 0) {\small $\mathit{d}_r$};
    \node[black!80] at (\cubexp-\cubex-0.3, \cubeyp-\cubey/2, \cubezp) {\small $\mathit{n_f}$};
    \node[black!80] at (\cubexp-\cubex/2  , \cubeyp-\cubey-0.3, \cubezp) {\small $\mathit{l_f}$};

%     \node[black] at (6.5, 0, 0) {$=$};
    \draw [->] (\cubexp+0.7, 0, 0) -- ++(0.5, 0, 0);
    
	% set of filters
    \pgfmathsetmacro{\cubexp}{8.4}
    \node[black] at (\cubexp-\cubex/2, \cubeyp+\cubey/2-0.2, \cubezp-\cubez/2) { $\mathbf{F}_r$};
    \draw[black,opacity=0.4,fill=black!60] (\cubexp,\cubeyp,0) -- ++(-\cubex,0,0) -- ++(0,-\cubey,0) -- ++(\cubex,0,0) -- cycle;
    \node[black!80] at (\cubexp-\cubex-0.3, \cubeyp-\cubey/2, 0) {\small $\mathit{n_f}$};
    \node[black!80] at (\cubexp-\cubex/2  , \cubeyp-\cubey-0.3, 0) {\small $\mathit{l_f}$};

    \draw [->] (\cubexp+0.4, 0, 0) -- ++(0.5, 0, 0);

	% implicit filter tensor
	\pgfmathsetmacro{\cubexp}{12.5}
    \pgfmathsetmacro{\cubezp}{1}
    \pgfmathsetmacro{\cubexx}{1.8}
    \node[black] at (\cubexp-\cubexx/2, \cubeyp+\cubey/2+0.2, \cubezp-\cubez/2) { $\mathcal{F}_r$};
    \draw[black,opacity=0.3,fill=black!60] (\cubexp,\cubeyp,\cubezp) -- ++(-\cubex,0,0) -- ++(0,-\cubey,0) -- ++(\cubex,0,0) -- cycle;
    \node[black] at (\cubexp-\cubex/2, \cubeyp-\cubey/2, \cubezp) { $\mathbf{F}_r$};
    
    \draw[black,opacity=0.25,fill=black!60] (\cubexp,\cubeyp,\cubezp) -- ++(-\cubexx+\cubex,0,-\cubez) -- ++(-\cubex,0,0) -- ++(\cubexx-\cubex,0,\cubez) -- cycle;
    \draw[black,opacity=0.25,fill=black!60] (\cubexp-\cubex,\cubeyp,\cubezp) -- ++(-\cubexx+\cubex,0,-\cubez) -- ++(0, -\cubey,0) -- ++(\cubexx-\cubex,0,\cubez) -- cycle;

\draw[black,opacity=0.15,fill=black!60] (\cubexp,\cubeyp,\cubezp) -- ++(-\cubexx,0,0) -- ++(0,-\cubey,0) -- ++(\cubexx,0,0) -- cycle;
    \draw[black,opacity=0.15,fill=black!60] (\cubexp,\cubeyp,\cubezp) -- ++(0,0,-\cubez) -- ++(0,-\cubey,0) -- ++(0,0,\cubez) -- cycle;
    \draw[black,opacity=0.15,fill=black!60] (\cubexp,\cubeyp,\cubezp) -- ++(-\cubexx,0,0) -- ++(0,0,-\cubez) -- ++(\cubexx,0,0) -- cycle;

	\node[black!80] at (\cubexp-\cubexx-0.2, \cubeyp+0.2, 0) {\small $\mathit{l_m}$};
    \node[black!80] at (\cubexp-\cubexx-0.3, \cubeyp-\cubey/2, \cubezp) {\small $\mathit{n_f}$};
    \node[black!80] at (\cubexp-\cubexx/2  , \cubeyp-\cubey-0.3, \cubezp) {\small $\mathit{d_e}$};

	% e_1
    \draw[black,opacity=0.8,fill=black!60] (\cubexp,\cubeyp-\cubey-1.5,\cubezp) -- ++(-\cubexx, 0, 0) -- ++(0, 0.3, 0) -- ++(+\cubexx, 0, 0) -- cycle;% e1
    \node[black] at (\cubexp-\cubexx/2,\cubeyp-\cubey-0.8, \cubezp) {$\mathbf{\otimes}_x$};
	\node[black] at (\cubexp-\cubexx/2,\cubeyp-\cubey-1.75, \cubezp) {$\mathbf{e}_1$};

    \node[black] at (\cubexp+1, \cubeyp-\cubey/2, 0) {$\mathbf{\otimes}_{yz}$};

	% W tensor
    \pgfmathsetmacro{\cubexp}{16.5}
    \node[black] at (\cubexp-\cubexx/2, \cubeyp+\cubey/2+0.2, \cubezp-\cubez/2) { $\mathcal{W}$};
    \draw[black,opacity=0.8,fill=black!60] (\cubexp,\cubeyp,\cubezp) -- ++(-\cubexx,0,0) -- ++(0,-\cubey,0) -- ++(\cubexx,0,0) -- cycle;
    \draw[black,opacity=0.8,fill=black!60] (\cubexp,\cubeyp,\cubezp) -- ++(0,0,-\cubez) -- ++(0,-\cubey,0) -- ++(0,0,\cubez) -- cycle;
    \draw[black,opacity=0.8,fill=black!60] (\cubexp,\cubeyp,\cubezp) -- ++(-\cubexx,0,0) -- ++(0,0,-\cubez) -- ++(\cubexx,0,0) -- cycle;
	\node[black!80] at (\cubexp-\cubexx-0.2, \cubeyp+0.2, 0) {\small $\mathit{l_m}$};
    \node[black!80] at (\cubexp-\cubexx-0.3, \cubeyp-\cubey/2, \cubezp) {\small $\mathit{n_f}$};
    \node[black!80] at (\cubexp-\cubexx/2  , \cubeyp-\cubey-0.3, \cubezp) {\small $\mathit{d}_e$};

	% RELU
    \draw [dashed] (\cubexp-\cubexx-.25,\cubeyp-\cubey-0.6, \cubezp) -- ++(\cubexx+.5, 0, 0);
    \node[black] at (\cubexp+0.5,\cubeyp-\cubey-0.6, \cubezp) {\textit{f}};

	% e_2	
    \node[black] at (\cubexp-\cubexx/2,\cubeyp-\cubey-0.85, \cubezp) {$\mathbf{\otimes}_x$};
    \draw[black,opacity=0.8,fill=black!60] (\cubexp,\cubeyp-\cubey-1.5,\cubezp) -- ++(-\cubexx, 0, 0) -- ++(0, 0.3, 0) -- ++(+\cubexx, 0, 0) -- cycle;% e1
	\node[black] at (\cubexp-\cubexx/2,\cubeyp-\cubey-1.75, \cubezp) {$\mathbf{e}_2$};

	% legend box
	\pgfmathsetmacro{\cubexl}{2.5}
    \pgfmathsetmacro{\cubeyl}{-1.4}
    \draw[-] (\cubexl,\cubeyl,\cubezp) -- ++(+6.55, 0, 0) -- ++(0, -1.45, 0) -- ++(-6.55, 0, 0) -- cycle;% e1
    
	% axes
	\pgfmathsetmacro{\cubexp}{\cubexl+0.3}
    \pgfmathsetmacro{\cubeyp}{\cubeyl-1.3}
	\draw [->, black] (\cubexp, \cubeyp, 0) -- ++(0.4, 0, 0);
    \draw [->, black] (\cubexp, \cubeyp, 0) -- ++(0, 0.4, 0);
    \draw [->, black] (\cubexp, \cubeyp, 0) -- ++(0, 0, 0.5);
	\node[black] at (\cubexp+0.6, \cubeyp    , 0  ) {\small $\mathit{x}$};
	\node[black] at (\cubexp    , \cubeyp+0.6, 0  ) {\small $\mathit{y}$};
	\node[black] at (\cubexp    , \cubeyp    , 0.8) {\small $\mathit{z}$};
    
	% legend
	\pgfmathsetmacro{\cubexp}{\cubexl+1.9}
    \pgfmathsetmacro{\cubeyp}{\cubeyl-0.45}
    \draw[black,opacity=0.8,fill=black!60] (\cubexp,\cubeyp,\cubezp) -- ++(-0.3, 0, 0) -- ++(0, 0.3, 0) -- ++(0.3, 0, 0) -- cycle;% e1
	\node[black, anchor=west] at (\cubexp,\cubeyp+0.15, \cubezp) {explicitly learned parameters};

    \pgfmathsetmacro{\cubeyp}{\cubeyl-0.9}
	\draw[black,opacity=0.4,fill=black!60] (\cubexp,\cubeyp,\cubezp) -- ++(-0.3, 0, 0) -- ++(0, 0.3, 0) -- ++(0.3, 0, 0) -- cycle;% e1
	\node[black, anchor=west] at (\cubexp,\cubeyp+0.15, \cubezp) {induced filter weights};

    \pgfmathsetmacro{\cubeyp}{\cubeyl-1.35}
	\draw[black,opacity=0.1,fill=black!60] (\cubexp,\cubeyp,\cubezp) -- ++(-0.3, 0, 0) -- ++(0, 0.3, 0) -- ++(0.3, 0, 0) -- cycle;% e1
	\node[black, anchor=west] at (\cubexp,\cubeyp+0.15, \cubezp) {zero weights};

  \end{tikzpicture}
}
  \caption{Interpretation of the HypER model in terms of tensor operations. Each relation embedding $\textbf{w}_r$ generates a set of filters $\mathbf{F}_r$ via the hypernetwork $\mathcal{H}$. The act of convolving $\mathbf{F}_r$ over $\mathbf{e}_1$ is equivalent to multiplication of $\mathbf{e}_1$ by a tensor $\mathcal{F}_r$ (in which $\mathbf{F}_r$ is diagonally duplicated and zero elsewhere). The tensor product $\mathcal{F}_r\otimes_{yz}\mathcal{W}$ gives a $d_e\times d_e$ matrix specific to each relation. Axes labels indicate the modes of tensor interaction (via inner product).}
  \label{fig:hypermatrix}
  \vspace{-0.7cm}
\end{figure}

\vspace{-0.3cm}
\subsection{Scoring Function and Model Architecture} \label{sec:score}

The relation-specific scoring function for the HypER model is:
\begin{equation}
\begin{split}
\phi_r(\mathbf{e}_1, \mathbf{e}_2) &= f(\text{vec}(\mathbf{e}_1 * \mathbf{F}_r)\mathbf{W})\mathbf{e}_2 \\
&= f(\text{vec}(\mathbf{e}_1 * \text{vec}^{-1}(\mathbf{w}_r\mathbf{H}))\mathbf{W})\mathbf{e}_2,
\end{split}
\end{equation}
where the $\text{vec}^{-1}$ operator reshapes a vector to a matrix, and non-linearity $f$ is chosen to be a rectified linear unit (ReLU).

In the feed-forward pass, the model obtains embeddings for the input triple from the entity and relation embedding matrices $\mathbf{E} \in \mathbb{R}^{n_e\times d_e}$ and $\mathbf{R} \in \mathbb{R}^{n_r\times d_r}$. The hypernetwork is a fully connected layer $\mathbf{H} \in \mathbb{R}^{d_r\times l_f n_f}$ ($l_f$ denotes filter length and $n_f$ the number of filters per relation, i.e. \textit{output channels} of the convolution) that is applied to the relation embedding $\mathbf{w}_r \in \mathbb{R}^{d_r}$. The result is reshaped to generate a matrix of convolutional filters $\mathbf{F}_r=\text{vec}^{-1}(\mathbf{w}_r\mathbf{H}) \in \mathbb{R}^{l_f \times n_f}$. Whilst the overall dimensionality of the filter set is $l_f n_f$, the rank is restricted to $d_r$ to encourage parameter sharing between relations. 

The subject entity embedding $\mathbf{e}_1$ is convolved with the set of relation-specific filters $\mathbf{F}_r$ to give a 2D feature map $\mathbf{M}_r \in \mathbb{R}^{l_m \times n_f}$, where $l_m = d_e - l_f +1$ is the feature map length. The feature map is vectorized to $\text{vec}(\mathbf{M}_r) \in \mathbb{R}^{l_m n_f}$, and projected to $d_e$-dimensional space by the weight matrix $\mathbf{W} \in \mathbb{R}^{l_m n_f\times d_e}$. After applying a ReLU activation function, the result is combined by way of inner product with each and every object entity embedding $\smash{{\mathbf{e}_2}^{(i)}}$, where $i$ varies over all entities in the dataset (of size $n_e$), to give a vector of scores. The logistic sigmoid is applied element-wise to the score vector to obtain the predicted probability of each prospective triple being true $\smash{\mathbf{p}_i = \sigma(\phi_r(\mathbf{e}_1, {\mathbf{e}_2}^{(i)}))}$.

\subsection{Understanding HypER as Tensor Factorization}

Having described the HypER architecture, we can view it as a series of tensor operations by considering the hypernetwork $\mathbf{H}$ and weight matrix $\mathbf{W}$ as tensors $\mathcal{H}\in \mathbb{R}^{d_r \times l_f \times n_f}$ and $\mathcal{W}\in \mathbb{R}^{l_m \times n_f \times d_e}$ respectively. The act of convolving $\mathbf{F}_r = \mathbf{w}_r\otimes\mathcal{H}$ over the subject entity embedding $\mathbf{e}_1$ is equivalent to the multiplication of $\mathbf{e}_1$ by a sparse tensor $\mathcal{F}_r$ within which $\mathbf{F}_r$ is diagonally duplicated with zeros elsewhere (see Figure \ref{fig:hypermatrix}). The result is multiplied by  $\mathcal{W}$ to give a vector, which is subject to ReLU before the final dot product with $\mathbf{e}_2$. Linearity allows the product $\mathcal{F}_r\otimes\mathcal{W}$ to be considered separately as generating a $d_e\times d_e$ matrix for each relation. Further, rather than duplicating entries of $\mathbf{F}_r$ within $\mathcal{F}_r$, we can generalize $\mathcal{F}_r$ to a relation-agnostic sparse 4 moded tensor $\mathcal{F}\in\mathbb{R}^{d_r \times d_e \times n_f \times l_m}$ by replacing entries with $d_r$-dimensional strands of $\mathcal{H}$. Thus, the HypER model can be described explicitly as tensor multiplication of $\mathbf{e}_1, \mathbf{e}_2$ and $\mathbf{w}_r$ with a core tensor $\mathcal{F}\otimes\mathcal{W}\in\mathbb{R}^{d_e\times d_e\times d_r}$, where $\mathcal{F}$ is heavily constrained in terms of its number of free variables. This insight allows HypER to be viewed in a very similar light to the family of factorization approaches to link prediction, such as RESCAL, DistMult and ComplEx.

\subsection{Training Procedure}

Following the training procedure introduced by \cite{dettmers2017convolutional}, we use \emph{1-N scoring} with the Adam optimizer \cite{kingma2014adam} to minimize the binary cross-entropy loss:
\begin{equation}
\mathcal{L}(\mathbf{p}, \mathbf{y}) = -\frac{1}{n_e}\sum_i (\mathbf{y}_i \text{log}(\mathbf{p}_i) + (1-\mathbf{y}_i) \text{log}(1-\mathbf{p}_i)),
\end{equation}
where $\mathbf{y} \in \mathbb{R}^{n_e}$ is the label vector containing ones for true triples and zeros otherwise, subject to \emph{label smoothing}.  \textbf{Label smoothing} is a widely used technique shown to improve generalization \cite{szegedy2016rethinking,pereyra2017regularizing}. Label smoothing changes the ground-truth label distribution by adding a uniform prior to encourage the model to be less confident, achieving a regularizing effect. \textbf{1-N scoring} refers to simultaneously scoring $(e_1, r, \mathcal{E})$, i.e. for all entities $e_2 \in \mathcal{E}$, in contrast to \emph{1-1 scoring}, the practice of training individual triples $(e_1, r, e_2)$ one at a time. As shown by \cite{dettmers2017convolutional}, 1-N scoring offers a significant speedup (3x on train and 300x on test time) and improved accuracy compared to 1-1 scoring. A potential extension of the HypER model described above would be to apply convolutional filters to \emph{both} subject and object entity embeddings. However, since this is not trivially implementable with 1-N scoring and wanting to keep its benefits, we leave this to future work. 

\subsection{Number of Parameters}

Table \ref{table:params} compares the number of parameters of ConvE and HypER (for the FB15k-237 dataset, which determines $n_e$ and $n_r$). It can be seen that, overall, HypER has fewer parameters (4.3M) than ConvE (5.1M) due to the way HypER directly transforms relations to convolutional filters.

\begin{table}[!htbp]
	\centering
	\caption{Comparison of number of parameters for ConvE and HypER on FB15k-237. $h_m$ and $w_m$ are height and width of the ConvE feature maps respectively.}
    \resizebox{7cm}{!}{
    \begin{tabular}{lcccc}
    \toprule 
    Model & $\mathbf{E}$ & $\mathbf{R}$ & Filters & $\mathbf{W}$\\
    \midrule 
    ConvE & \parbox{1.3cm}{\centering$n_e \times d_e\newline2.9M$} & \parbox{1.3cm}{\centering$ n_r \times d_r\newline0.1M$} & \parbox{1.3cm}{\centering$\phantom{0}l_f n_f\newline0.0M$} & \parbox{2.5cm}{\centering$h_m w_m n_f\times d_e\newline2.1M$} \vspace{0.2cm}\\
    HypER & \parbox{1.3cm}{\centering$ n_e \times d_e\newline2.9M$} & \parbox{1.3cm}{\centering$ n_r \times d_r\newline0.1M$} & \parbox{1.3cm}{\centering$d_r\times l_f n_f\newline0.1M$} & \parbox{2.5cm}{\centering$l_m n_f\times d_e\newline1.2M$} \vspace{0.1cm}\\
    \bottomrule
    \end{tabular}
    }
     \label{table:params}
      \vspace{-0.3cm}
 \end{table}

\section{Experiments}

\subsection{Datasets}

We evaluate our HypER model on the standard link prediction task using the following datasets (see Table \ref{table:datasets}):

\keypoint{FB15k}  \cite{bordes2013translating} a subset of Freebase, a large database of facts about the real world.

\keypoint{WN18}  \cite{bordes2013translating} a subset of WordNet, containing lexical relations between words.

\keypoint{FB15k-237} created by \cite{toutanova2015representing}, noting that the validation and test sets of FB15k and WN18 contain the inverse of many relations present in the training set, making it easy for simple models to do well. FB15k-237 is a subset of FB15k with the inverse relations removed.

\keypoint{WN18RR} \cite{dettmers2017convolutional} a subset of WN18, created by removing the inverse relations. 

\keypoint{YAGO3-10} \cite{dettmers2017convolutional} a subset of YAGO3 \cite{mahdisoltani2013yago3}, containing entities which have a minimum of 10 relations each. 
\vspace{-0.5cm}
\begin{table}[!htp]
	\centering
	\caption{Summary of dataset statistics.}
    \resizebox{5.3cm}{!}{
      \begin{tabular}{lrr}
        \toprule 
        Dataset & \qquad Entities ($n_e$) & \ \ Relations ($n_r$)\\
        \midrule 
        FB15k  		&	14,951 	& 1,345 \\
        WN18 		& 	40,943 	& 18 	\\
        FB15k-237	&	14,541	& 237	\\ 
        WN18RR		& 	40,943 	& 11	\\
        YAGO3-10	& 	123,182 & 37 	\\
        \bottomrule
      \end{tabular}
    }
     \label{table:datasets}
 \end{table}

\vspace{-1cm}
\subsection{Experimental Setup}

We implement HypER in PyTorch \cite{paszke2017automatic} and make our code publicly available.\footnote{\texttt{https://github.com/ibalazevic/HypER}}

\keypoint{Implementation Details}
We train our model with 200 dimension entity and relation embeddings ($d_e=d_r=200$) and 1-N scoring. Whilst the relation embedding dimension does not have to equal the entity embedding dimension, we set $d_r=200$ to match ConvE for fairness of comparison.

To accelerate training and prevent overfitting, we use batch normalization \cite{ioffe2015batch} and dropout \cite{srivastava2014dropout} on the input embeddings, feature maps and the hidden layer. We perform a hyperparameter search and select the best performing model by mean reciprocal rank (MRR) on the validation set. Having tested the values $\{0., 0.1, 0.2, 0.3\}$, we find that the following combination of parameters works well across all datasets: input dropout 0.2, feature map dropout 0.2, and hidden dropout 0.3, apart from FB15k-237, where we set input dropout to 0.3. We select the learning rate from $\{0.01, 0.005, 0.003, 0.001, 0.0005, 0.0001\}$ and exponential learning rate decay from $\{1., 0.99, 0.995\}$ for each dataset and find the best performing learning rate and learning rate decay to be dataset-specific. We set the convolution stride to 1, number of feature maps to 32 with the filter size $3\times3$ for ConvE and $1 \times 9$ for HypER, after testing different numbers of feature maps $ n_f \in \{16, 32, 64\}$ and filter sizes $l_f \in \{1\times1, 1\times2, 1\times3, 1\times6, 1\times9, 1\times12\}$ (see Table \ref{table:ablation}). We train all models using the Adam optimizer with batch size 128. One epoch on FB15k-237 takes approximately 12 seconds on a single GPU compared to 1 minute for e.g. RESCAL, largely due to 1-N scoring. 

\keypoint{Evaluation} Results are obtained by iterating over all triples in the test set. A particular triple is evaluated by replacing the object entity $e_2$ with all entities $\mathcal{E}$ while keeping the subject entity $e_1$ fixed and vice versa, obtaining scores for each combination. These scores are then ranked using the ``filtered'' setting only, i.e. we remove all true cases other than the current test triple \cite{bordes2013translating}.  

We evaluate HypER on five different metrics found throughout the link prediction literature: mean rank (MR), mean reciprocal rank (MRR), hits@10, hits@3, and hits@1. Mean rank is the average rank assigned to the true triple, over all test triples. Mean reciprocal rank takes the average of the reciprocal rank assigned to the true triple. Hits@k measures the percentage of cases in which the true triple appears in the top k ranked triples. Overall, the aim is to achieve high mean reciprocal rank and hits@k and low mean rank. For a more extensive description of how each of these metrics is calculated, we refer to \cite{dettmers2017convolutional}.

\subsection{Results}
 Link prediction results for all models across the five datasets are shown in Tables \ref{table:wn18rr}, \ref{table:wn18} and \ref{table:yago}. Our key findings are:

\begin{itemize}
\item whilst having fewer parameters than the closest comparator ConvE, HypER consistently outperforms all other models across all datasets, thereby achieving state-of-the-art results on the link prediction task; and
\item our filter dimension study suggests that no benefit is gained by convolving over reshaped 2D entity embeddings in comparison with 1D entity embedding vectors and that most information can be extracted with very small convolutional filters (Table \ref{table:ablation}).
\end{itemize}

Overall, HypER outperforms all other models on all metrics apart from mean reciprocal rank on WN18 and mean rank on WN18RR, FB15k-237, WN18, and YAGO3-10. Given that mean rank is known to be highly sensitive to outliers \cite{nickel2016holographic}, this suggests that HypER correctly ranks many true triples in the top 10, but makes larger ranking errors elsewhere. 

Given that most models in the literature, with the exception of ConvE, were trained with 100 dimension embeddings and 1-1 scoring, we reimplement previous models (DistMult, ComplEx and ConvE) with 200 dimension embeddings and 1-N scoring for fair comparison and report the obtained results on WN18RR in Table \ref{table:wn18rr_our}. We perform the same hyperparameter search for every model and present the mean and standard deviation of each result across five runs (different random seeds). This improves most previously published results, except for ConvE where we fail to replicate some values. Notwithstanding, HypER remains the best performing model overall despite better tuning of the competitors.

\begin{table}[!htbp]
\vspace{-0.4cm}
	\centering
	\caption{Link prediction results on WN18RR and FB15k-237. The RotatE \cite{sun2019rotate} results are reported without their self-adversarial negative sampling (see Appendix H in the original paper) for fair comparison, given that it is not specific to that model only.}
    \resizebox{8cm}{!}{
    \begin{tabular}{lccccccccccc}
    \toprule 
    &\multicolumn{5}{c}{WN18RR}&&\multicolumn{5}{c}{FB15k-237}\\ 
    \cmidrule{2-6} \cmidrule{8-12}
    & MR & MRR & H@10 & H@3 & H@1 & & MR & MRR & H@10 & H@3 & H@1\\
    \midrule
	DistMult \cite{yang2014embedding} & $5110$ & $.430$ & $.490$ & $.440$ & $.390$ & & $254$ & $.241$ & $.419$ & $.263$ & $.155$\\    
    ComplEx \cite{trouillon2016complex} & $5261$ & $.440$ & $.510$ & $.460$ & $.410$ & & $339$ & $.247$ & $.428$ & $.275$ & $.158$ \\ 
    Neural LP \cite{yang2017differentiable} & $-$ & $-$ & $-$ & $-$ & $-$ & & $-$ & $.250$ & $.408$ & $-$ & $-$ \\
    R-GCN \cite{schlichtkrull2018modeling} & $-$ & $-$ & $-$ & $-$ & $-$ & & $-$ & $.248$ & $.417$ & $.264$ & $.151$ \\
    MINERVA \cite{das2018go} & $-$ & $-$ & $-$ & $-$ & $-$ & & $-$ & $-$ & $.456$ & $-$ & $-$\\
    ConvE \cite{dettmers2017convolutional} & $\mathbf{4187}$ & $.430$ & $.520$ & $.440$ & $.400$ & & $244$ & $.325$ & $.501$ & $.356$ & $.237$ \\ 
    M-Walk \cite{shen2018m} & $-$ & $.437$ & $-$ & $.445$ & $.414$ & & $-$ & $-$ & $-$ & $-$ & $-$ \\ 
    RotatE \cite{sun2019rotate} & $-$ & $-$ & $-$ & $-$ & $-$ & & $\mathbf{185}$ & $.297$ & $.480$ & $.328$ & $.205$\\
    \midrule
    HypER (ours) & $5798$ & $\mathbf{.465}$ & $\mathbf{.522}$ & $\mathbf{.477}$ & $\mathbf{.436}$ & & $250$ & $\mathbf{.341}$ &$\mathbf{.520}$& $\mathbf{.376}$ & $\mathbf{.252}$\\
    \bottomrule
    \end{tabular}
    }
     \label{table:wn18rr}
 \end{table}

 \begin{table}[!htbp]
   \vspace{-1cm}
	\centering
	\caption{Link prediction results on WN18 and FB15k.}
    \resizebox{8cm}{!}{
    \begin{tabular}{lccccccccccc}
    \toprule 
    &\multicolumn{5}{c}{WN18}&&\multicolumn{5}{c}{FB15k}\\ 
    \cmidrule{2-6} \cmidrule{8-12}
    & MR & MRR & H@10 & H@3 & H@1 & & MR & MRR & H@10 & H@3 & H@1\\
    \midrule
    TransE \cite{bordes2013translating} & $\mathbf{251}$ & $-$ & $.892$ & $-$ & $-$ & & $125$ & $-$ & $.471$ & $-$ & $-$\\ 
	DistMult \cite{yang2014embedding} & $902$ & $.822$ & $.936$ & $.914$ & $.728$ & & $97$ & $.654$ & $.824$ & $.733$ & $.546$\\    
    ComplEx \cite{trouillon2016complex} & $-$ & $.941$ & $.947$ & $.936$ & $.936$ & & $-$ & $.692$ & $.840$ & $.759$ & $.599$ \\ 
    ANALOGY \cite{liu2017analogical} & $-$ & $.942$ & $.947$ & $.944$ & $.939$ & & $-$ & $.725$ & $.854$ & $.785$ & $.646$ \\ 
    Neural LP \cite{yang2017differentiable} & $-$ & $.940$ & $.945$ & $-$ & $-$ & & $-$ & $.760$ & $.837$ & $-$ & $-$ \\
    R-GCN \cite{schlichtkrull2018modeling} & $-$ & $.819$ & $\mathbf{.964}$ & $.929$ & $.697$ & & $-$ & $.696$ & $.842$ & $.760$ & $.601$ \\
    TorusE \cite{ebisu2018toruse} & $-$ & $.947$ & $.954$ & $.950$ & $.943$ & & $-$ & $.733$ & $.832$ & $.771$ & $.674$ \\
    ConvE \cite{dettmers2017convolutional} & $374$ & $.943$ & $.956$ & $.946$ & $.935$ & & $51$ & $.657$ & $.831$ & $.723$ & $.558$ \\ 
    SimplE \cite{kazemi2018simple} & $-$ & $.942$ & $.947$ & $.944$ & $.939$ & & $-$ & $.727$ & $.838$ & $.773$ & $.660$ \\ 
    \midrule
    HypER (ours) & $431$ & $\mathbf{.951}$ & $.958$ & $\mathbf{.955}$ & $\mathbf{.947}$ & & $\mathbf{44}$ & $\mathbf{.790}$ &$\mathbf{.885}$& $\mathbf{.829}$ & $\mathbf{.734}$\\
    \bottomrule
    \end{tabular}
    }
     \label{table:wn18}
 \end{table}
\begin{table}[!htbp]
 \vspace{-1cm}
	\centering
	\caption{Link prediction results on YAGO3-10.}
    \resizebox{5cm}{!}{
    \begin{tabular}{lccccc}
    \toprule 
    &\multicolumn{5}{c}{YAGO3-10}\\ 
    \cmidrule{2-6}
     & MR & MRR & H@10 & H@3 & H@1\\
    \midrule
	 DistMult \cite{yang2014embedding} & $5926$ & $.340$ & $.540$ & $.380$ & $.240$\\
     ComplEx \cite{trouillon2016complex} & $6351$ & $.360$ & $.550$ & $.400$ & $.260$\\
     ConvE \cite{dettmers2017convolutional} & $\mathbf{1676}$ & $.440$ & $.620$ & $.490$ & $.350$\\
    \midrule
    HypER (ours) & $2529$ & $\mathbf{.533}$ & $\mathbf{.678}$ & $\mathbf{.580}$ & $\mathbf{.455}$\\
    \bottomrule
    \end{tabular}
    }
     \label{table:yago}
      \vspace{-0.4cm}
 \end{table}

\begin{table}[!htbp]
	\centering
	 \caption{Link prediction results on WN18RR; all models trained with 200 dimension embeddings and 1-N scoring.}
    \resizebox{9cm}{!}{
    \begin{tabular}{lccccc}
    \hline 
    &\multicolumn{5}{c}{WN18RR}\\ 
    \cline{2-6}
    & MR & MRR & H@10 & H@3 & H@1\\
    \hline
	DistMult \cite{yang2014embedding}& $\mathbf{4911 \pm 109}$ & $.434 \pm .002$ & $.508 \pm .002$ & $.447 \pm .001$ & $.399 \pm .002$\\    
    ComplEx \cite{trouillon2016complex} & $5930 \pm 125$ & $.446 \pm .001$ & $\mathbf{.523 \pm .002}$ & $.462 \pm .001$ & $.409 \pm .001$ \\ 
    ConvE \cite{dettmers2017convolutional} & $\mathbf{4997 \pm}$\phantom{0}$\mathbf{99}$ & $.431 \pm .001$ & $.504 \pm .002$ & $.443 \pm .002$ & $.396 \pm .001$ \\ 
    \hline
    HypER (ours) & $5798 \pm 124$ & $\mathbf{.465 \pm .002}$ & $\mathbf{.522 \pm .003}$ & $\mathbf{.477 \pm .002}$ & $\mathbf{.436 \pm .003}$ \\
    \hline
    \end{tabular}
    }
     \label{table:wn18rr_our}
     \vspace{-0.2cm}
 \end{table}

 To ensure that the difference between reported results for HypER and ConvE is not simply due to HypER having a reduced number of parameters (implicit regularization), we trained ConvE reducing the number of feature maps to 16 instead of 32 to have a comparable number of parameters to HypER (explicit regularization). This showed no improvement in ConvE results, indicating HypER's architecture does more than merely reducing the number of parameters.

\begin{table}[!ht]
	\centering
	\caption{Results with and without hypernetwork on WN18RR and FB15k-237.}
    \resizebox{8cm}{!}{
    \begin{tabular}{lccccc}
    \toprule 
    &\multicolumn{2}{c}{WN18RR}&&\multicolumn{2}{c}{FB15k-237}\\ 
    \cmidrule{2-3} \cmidrule{4-6}
    & MRR & H@10 & & MRR & H@10\\
    \midrule
    HypER & $\mathbf{.465 \pm .002}$ & $\mathbf{.522 \pm .003}$ & & $\mathbf{.341 \pm .001}$ &$\mathbf{.520 \pm .002}$\\
    HypER (no $\mathbf{H}$) & $.459 \pm .002$ & $.511 \pm .002$ & & $.338 \pm .001$ &$.515 \pm .001$\\ 
    \bottomrule≥
    \end{tabular}
    }
    \label{table:without_hyper}
    \vspace{-0.8cm}
 \end{table}
\keypoint{Hypernetwork Influence} To test the influence of the hypernetwork and, thereby, knowledge sharing between relations, we compare HypER results on WN18RR and FB15k-237 with the hypernetwork component removed, i.e. without the first fully connected layer and with the relation embeddings directly corresponding to a set of convolutional filters. Results presented in Table \ref{table:without_hyper} show that the hypernetwork component improves performance, demonstrating the value of multi-task learning across different relations.

\keypoint{Filter Dimension Study} Table \ref{table:ablation} shows results of our study investigating the influence of different convolutional filter sizes on the performance of HypER. The lower part of the table shows results for 2D filters convolved over reshaped ($10 \times 20$) 2D subject entity embeddings. It can be seen that reshaping the embeddings is of no benefit, especially on WN18RR. These results indicate that the purpose of convolution on word embeddings is not to find patterns in a 2D embedding (as with images), but perhaps to limit the number of dimensions that can interact with each other, thereby avoiding overfitting. In the upper part of the table, we vary the length of 1D filters, showing that comparable results can be achieved with filter sizes $1 \times 6$ and $1 \times 9$, with diminishing results for smaller (e.g. $1 \times 1$) and larger (e.g. $1 \times 12$) filters.

\begin{table}[!htbp]
	\centering
	\caption{Influence of different filter dimension choices on prediction results.}
    \resizebox{4cm}{!}{
    \begin{tabular}{cccccc}
    \toprule 
    &\multicolumn{2}{c}{WN18RR} & &\multicolumn{2}{c}{FB15k-237}\\ 
    \cmidrule{2-3}\cmidrule{5-6}
    Filter Size & MRR & H@1 & & MRR & H@1\\
    \midrule
	 $1 \times 1$ & $.455$ & $.422$ & & $.337$ & $.248$\\   
     $1 \times 2$ & $.458$ & $.428$ & & $.337$ & $.248$ \\
     $1 \times 3$ & $.457$ & $.427$ & & $.339$ & $.250$ \\   
     $1 \times 6$ & $.459$ & $.429$ & & $.340$ & $.251$ \\   
     $1 \times 9$ & $\mathbf{.465}$ & $\mathbf{.436}$ & & $\mathbf{.341}$ & $\mathbf{.252}$\\  
     $1 \times 12$ & $.457$ & $.428$ & & $\mathbf{.341}$ & $\mathbf{.252}$\\   
    \midrule
    $2 \times 2$ & $.456$ & $.429$ & & $.340$ & $.250$\\
    $3 \times 3$ & $.458$ & $.430$ & & $.339$ & $.250$\\
    $5 \times 5$ & $.452$ & $.423$ & & $.340$ & $\mathbf{.252}$\\
    \bottomrule
    \end{tabular}
    }
     \label{table:ablation}
 \end{table}

\keypoint{Label Smoothing}Contrary to the ablation study of \cite{dettmers2017convolutional}, showing the influence of hyperparameters on mean reciprocal rank for FB15k-237, from which they deem label smoothing unimportant, we find label smoothing to give a significant improvement in prediction scores for WN18RR. However, we find it does have a negative influence on the FB15k scores and as such, exclude label smoothing from our experiments on that dataset. We therefore recommend evaluating the influence of label smoothing on a per dataset basis and leave to future work analysis of the utility of label smoothing in the general case.

\section{Conclusion}
In this work, we introduce HypER, a hypernetwork model for link prediction on knowledge graphs. HypER generates relation-specific convolutional filters and applies them to subject entity embeddings. The hypernetwork component allows information to be shared between relation vectors, enabling multi-task learning across relations. To our knowledge, HypER is the first link prediction model that creates  non-linear interaction between entity and relation embeddings by convolving relation-specific filters over the entity embeddings. 

We  show that no benefit is gained from 2D convolutional filters over 1D, dispelling the suggestion that 2D structure exists in entity embeddings implied by ConvE. We also recast HypER in terms of tensor operations showing that, despite the convolution operation, it is closely related to the established family of tensor factorization models. Our results suggest that convolution provides a good trade-off between expressiveness and parameter number compared to a dense network. HypER is fast, robust to overfitting, has relatively few parameters, and achieves state-of-the-art results across almost all metrics on multiple link prediction datasets.

Future work might include expanding the current architecture by applying convolutional filters to both subject and object entity embeddings. We may also analyze the influence of label smoothing and explore the interpretability of convolutional feature maps to gain insight and potentially improve the model.

\subsection*{Acknowledgements}
We thank Ivan Titov for helpful discussions on this work. Ivana Bala\v{z}evi\'c and Carl Allen were supported by the Centre for Doctoral Training in Data Science, funded by EPSRC (grant EP/L016427/1) and the University of Edinburgh.

\bibliographystyle{splncs04}
\bibliography{hyper}

\end{document}